\documentclass{article}

\usepackage[nonatbib, preprint]{neurips_2023}

\usepackage[utf8]{inputenc} 
\usepackage[T1]{fontenc}    
\usepackage{hyperref}       
\usepackage{url}            
\usepackage{booktabs}       
\usepackage{amsfonts}       
\usepackage{nicefrac}       
\usepackage{microtype}      
\usepackage{xcolor}         
\usepackage{graphicx}
\usepackage{amsmath}

\title{Compressing Vision Transformers for Low-Resource Visual Learning}

\author{
  Eric Youn \\
  \And
  Sai Mitheran J \\
  \And
  Sanjana Prabhu \\
  \And
  Siyuan Chen  
  \And
   \textbf{Carnegie Mellon University} \\
  \texttt{\{eyoun, sjagades, sprabhu2, siyuanch\}@andrew.cmu.edu} \\
}

\begin{document}

\maketitle


\section{Motivation}
\label{sec:motivation}
Vision transformer (ViT) and its variants have swept through visual learning leaderboards and offer state-of-the-art accuracy in tasks such as image classification, object detection, and semantic segmentation by attending to different parts of the visual input and capturing long-range spatial dependencies. However, these models are large and computation-heavy. For instance, the recently proposed ViT-B~\cite{dosovitskiy2021image} model has 86M parameters making it impractical for deployment on resource-constrained devices. As a result, their deployment on mobile and edge scenarios is limited. In our work, we aim to take a step toward bringing vision transformers to the edge by utilizing popular model compression techniques such as distillation, pruning, and quantization. 

Our chosen application environment is an unmanned aerial vehicle (UAV) that is battery-powered and memory-constrained, carrying a single-board computer on the scale of an NVIDIA Jetson Nano with 4GB of RAM. On the other hand, the UAV requires high accuracy close to that of state-of-the-art ViTs to ensure safe object avoidance in autonomous navigation, or correct localization of humans in search-and-rescue. Inference latency should also be minimized given the application requirements. Hence, our target is to enable rapid inference of a vision transformer on an NVIDIA Jetson Nano (4GB) with minimal accuracy loss. This allows us to deploy ViTs on resource-constrained devices, opening up new possibilities in surveillance, environmental monitoring, etc. Our implementation is made available at \href{https://github.com/chensy7/efficient-vit}{https://github.com/chensy7/efficient-vit}.

\section{Related Work}
\label{sec:related}
\textbf{Knowledge distillation} is a technique that improves the performance of a smaller student model by using "soft" labels from a larger teacher model: this does not compress the model in any dimension of the network~\cite{hinton2015distilling}. Soft labels are thought to be more informative than hard labels, which can benefit student training. Soft labels effectively transfer the knowledge of the teacher to the student model~\cite{yuan2020revisiting}.

\textbf{Pruning} techniques are classified into two types: unstructured~\cite{lee2018snip} and structured~\cite{luo2017thinet}. Unstructured pruning entails removing insignificant weights based on specific criteria, resulting in irregular sparsity that makes hardware acceleration difficult. Structured pruning addresses this by assigning a weight to groups of parameters, such as convolutional channels or matrix rows~\cite{lin2018accelerating}. Sensitivity can be determined using a variety of metrics, including batch normalization layer scaling factors, channel-wise summation over weights, channel rank, and the geometric median of convolutional filters. Common structures to prune in transformer-based models include blocks, attention heads, and fully-connected matrix rows~\cite{https://doi.org/10.48550/arxiv.2106.04533}. Several studies have proposed various methods for pruning redundant heads, entire layers, and extracting shallow models during inference~\cite{https://doi.org/10.48550/arxiv.1905.10650}.

\textbf{Quantization}-Aware Training (QAT)~\cite{zhou2016dorefa} and Post-Training Quantization (PTQ)~\cite{https://doi.org/10.48550/arxiv.1810.05723} are the two most common quantization methods. To achieve highly compressed quantization, QAT relies on training, which frequently necessitates significant expert knowledge and GPU resources for training or fine-tuning. PTQ has gained popularity as a way to reduce the costs associated with quantization since additional training is not required. Many excellent works have resulted from PTQ, including OMSE~\cite{choukroun2019low}, which reduces quantization errors by determining the value range of activation, and AdaRound~\cite{nagel2020adaround}, which adapts data and task loss via a novel rounding mechanism. Liu et al.~\cite{liu2021post} proposed a post-training quantization method for vision transformers that employs similarity-aware and rank-aware strategies, but it excludes Softmax and LayerNorm modules, resulting in incomplete quantization.

 One promising direction to reduce the computational complexity of the self-attention mechanism in Vision Transformers (ViT) while maintaining or even improving their performance is the use of \textbf{low-rank approximation} of attention matrices. This can significantly reduce the number of computations needed for each self-attention operation. Several studies have investigated different variants of low-rank attention mechanisms, such as structured and unstructured low-rank approximations, tensor decompositions, and dynamic rank selection~\cite{lee2020parameter, chen2021scatterbrain, dass2022vitality, qin2022dba}. These methods have been shown to be effective in reducing the computational cost of ViT while achieving competitive performance on various image classification benchmarks.

\textbf{Efficient/sublinear-regime attention} mechanisms in the Natural Language Processing domain attempts to reduce the quadratic complexity of standard Multi-Head Self-Attention (MSA). To accomplish this, various methods have been developed, including low-rank decomposition~\cite{https://doi.org/10.48550/arxiv.2006.04768}, kernelization~\cite{https://doi.org/10.48550/arxiv.2006.16236}, memory, and sparsity mechanisms~\cite{https://doi.org/10.48550/arxiv.1904.10509}. These methods, however, have shown drawbacks in Computer Vision (CV) tasks, where Spatial Reduction Attention (SRA)~\cite{https://doi.org/10.48550/arxiv.2102.12122} and Local Windowed Attention~\cite{https://doi.org/10.48550/arxiv.2103.14030} are used instead. However, these mechanisms only pay attention to either local or global issues. Some approaches addressed this limitation by introducing additional global tokens or combining local window attention with depthwise convolutional layers~\cite{https://doi.org/10.48550/arxiv.2204.02557}. Other attention mechanisms consider both local and global attention~\cite{https://doi.org/10.48550/arxiv.2201.02767}, but due to inefficient operations, they may be slow on GPUs.

\begin{figure*}[!h]
    \centering
    \includegraphics[scale=0.45]{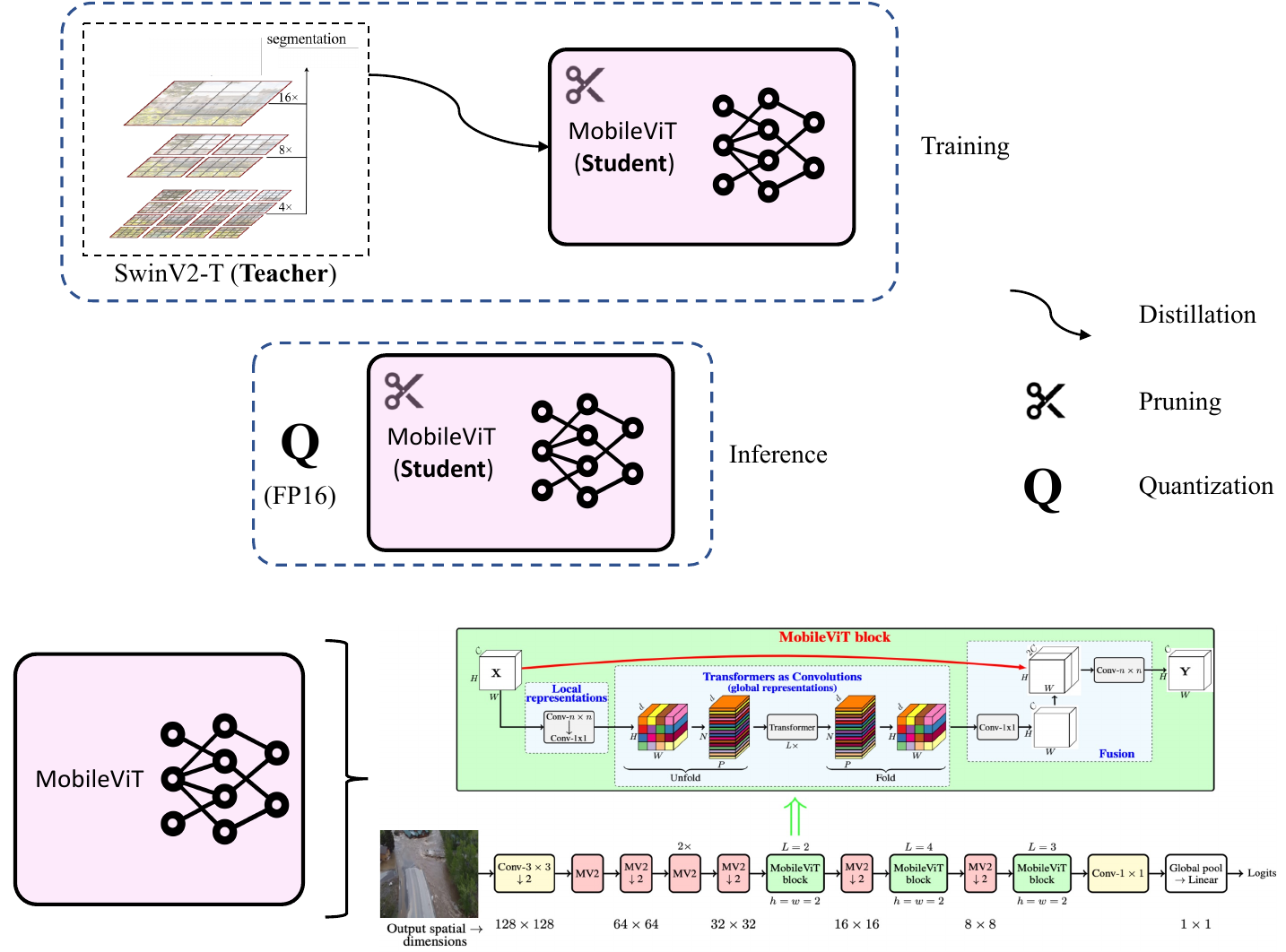} 
    \caption{Our framework for edge-oriented, rapid on-device disaster scene parsing}
    \label{fig:model}
\end{figure*}

\section{Methodology}
\label{sec:method}

\subsection{Background and Formulation}
    To meet the requirements of utilizing no more than 4GB of RAM system-wide, we incorporate most of the above approaches to achieve a compressed model with Swin Transformer level accuracy as shown in Figure~\ref{fig:model}. In order to balance both memory usage and inference latency improvements, we propose the use of structured pruning to remove redundant heads from the attention mechanism and neurons from the linear layers.~\cite{https://doi.org/10.48550/arxiv.2209.02432} observed that many traditional distillation techniques from CNNs do not carry over to vision transformers. They propose a mimicking and generation-based distillation approach. We utilize a similar approach to our distillation framework, using logit knowledge transfer at the prediction stage, coupled with mimicking feature distillation at early layers and generation-based feature distillation in later layers. Our pruning technique focuses on convolution layers, linear layers, and entire heads, but the actual attention matrices are not compressed. Therefore, we can potentially build on this and use low-rank approximation to further compress the model. Our final avenue for compression leverages post-training quantization which is quite limited in PyTorch due to the lack of robust support for quantized CUDA operations. Nevertheless, we quantize our model to the maximum possible extent to reduce memory requirements while keeping the performance at level.

Our target is ambitious - a tiny Swin Transformer has at least 40M parameters while we estimate the maximum parameter size for a model running on the Jetson Nano 4GB is about 5M. However, by applying the approaches specified above, we push the accuracy of the model deployed on the edge device as close as possible to the state-of-the-art level (Swin Transformer). Our task involves segmentation of search-and-rescue categories and therefore is more specialized and easier than datasets such as ADE20K or COCO. Therefore, a smaller network will intuitively be able to capture the information across these classes.

\subsection{Relation to prior work}
As mentioned in the previous sections, much of the knowledge for model compression in previous architectures and NLP-based transformers does not translate well to vision transformers. Additionally, much of the prior work in this area focuses only on certain compression techniques rather than unified models combining various techniques together. One notable exception is the work by Yu et al.~\cite{https://doi.org/10.48550/arxiv.2203.08243} which proposes a unified framework using a structured pruning approach to compress the attention module, knowledge distillation, and using skip connections as a mechanism to drop entire blocks. However, they do not include quantization and only conduct distillation at the prediction logits stage. Similarly, certain works~\cite{https://doi.org/10.48550/arxiv.2203.13444} focus on a combined structured pruning and low-rank approximation framework but do not include distillation for further compression. Our contribution to this line of research is to combine modern techniques spread throughout the literature into a \textit{unified compression framework} and apply this approach to an ambitious real-world problem: our on-device compression framework is applied for disaster scene parsing, where the goal is to segment and parse images of UAV-view disaster scenes using a model that runs efficiently on a Jetson Nano device. We test our compression framework by performing rapid inference using the transformer model on an NVIDIA Jetson Nano 4GB, finding the sweet spot between performance and compute requirements.

\subsection{Dataset}

\begin{figure*}[!h]
    \centering
    \includegraphics[scale=0.3]{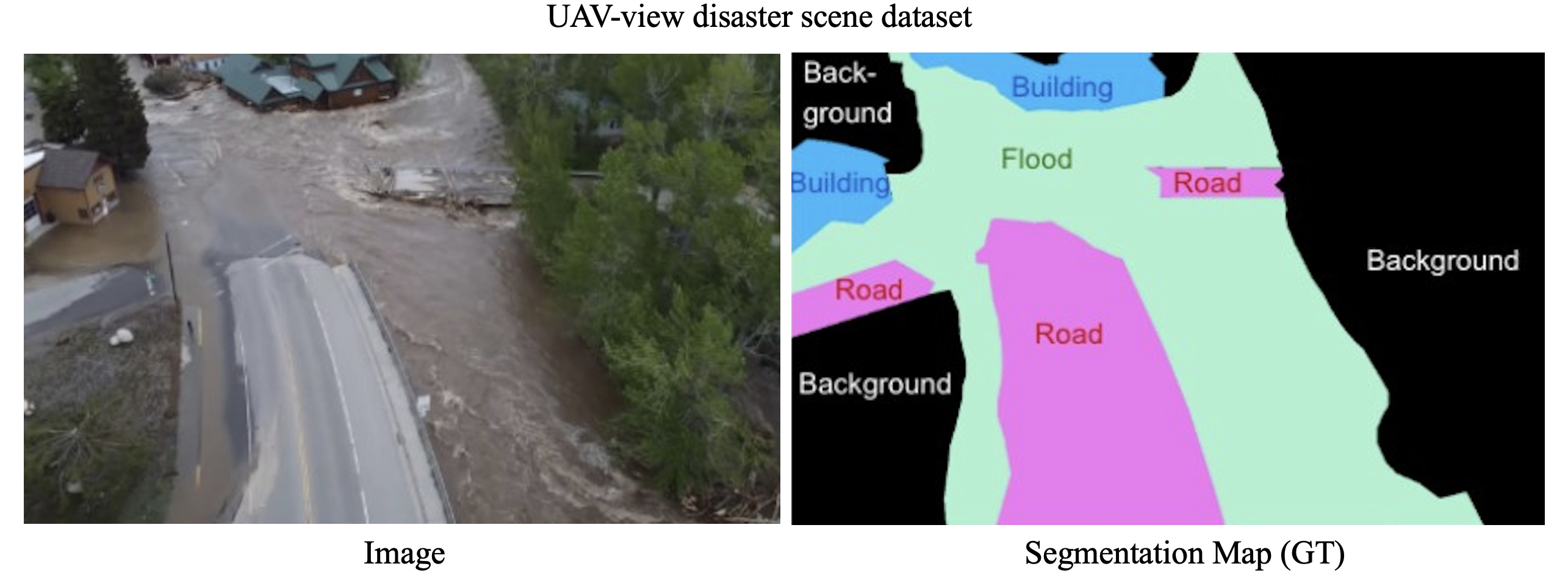} 
    \caption{Samples from the LPCV dataset (CVPR 2023 challenge)}
    \label{fig:cif10}
\end{figure*}

Our work utilizes the LPCV 2023 challenge dataset for the on-device disaster scene parsing task. This dataset, which is already publicly available online, is a UAV-view disaster scene dataset consisting of 1,720 images collected from various disaster scenes.

Out of the total images, we use only 1,120 images along with their ground-truth labels for training our model and testing its performance, while the remaining images are used for benchmarking submissions on the challenge server. The images are densely annotated with 14 categories, including \textit{Background, Avalanche, Building\_undamaged, Building\_damaged, Cracks/fissure/subsidence, Debris/mud/rock\_flow, Fire/flare, Flood/water/river/sea, Ice\_jam\_flow, Lava\_flow, Person, Pyroclastic\_flow, Road/railway/bridge, and Vehicle}, all resized to a resolution of 512x512 pixels.

Our use of this dataset will aid in the development of more accurate and efficient disaster response systems and contribute to the existing body of research in the field of building efficient, performant systems for on-the-edge use cases, such as in the case of natural disasters and remote area access.

\subsection{Evaluation Metrics}
Our work is inspired by the 2023 LPCV Challenge), for CVPR. Our target task is dense prediction, i.e., scene parsing or semantic segmentation. Following the challenge, we plan to evaluate our framework using the following metrics: for \textbf{accuracy}, we use \textit{mean IoU}, as shown below ($\mathrm{n}$ is the total number of test images; TP, FP, and FN correspond to the True Positive, False Positive, and False Negative pixels). Then, \textbf{execution time} is measured as the average inference time over all test images. Our \textbf{final score} is computed as accuracy per latency, as shown in the following formulae.

\begin{align}
        \text { mean IoU } & =\frac{\sum_{i=1}^n\left(\frac{T P_i}{T P_i+F N_i+F P_i}\right)}{n} 
\end{align}

\begin{align}
        \text { Final Score } & =\frac{\text { mean IoU }}{\text { Execution Time }}
\end{align}

In addition, we measure \textbf{model size} given general memory constraints on the edge. Although we have set a clear upper bound for memory (4GB), these metrics tell us if our solution can be extended to even more resource-constrained scenarios.

Furthermore, we conduct an ablation study to understand how each compression method (distillation, pruning, quantization) contributes to the accuracy versus latency tradeoff.






\subsection{Dataset and Software}
We use PyTorch for training and inference procedures. While faster inference engines are available, they are mostly orthogonal to our solution. We use the LPCV dataset for knowledge distillation, quantization-aware training, and finetuning after pruning. Our implementation is made available at \href{https://github.com/chensy7/efficient-vit}{https://github.com/chensy7/efficient-vit}.

\subsection{Baselines}
\label{sec:baselines}
We use Swin Transformer~\cite{https://doi.org/10.48550/arxiv.2103.14030} and MobileViT~\cite{Mehta2021MobileViTLG} as our primary encoder backbones. The Swin Transformer is well-designed with unique features such as the use of shifted windows to compute the representation, which allows it to capture more global information while maintaining local details. MobileViT is tailored to be used in mobile devices that have limited memory and computational resources. It accomplishes this by replacing fully connected layers with convolutional layers in the MLP blocks of the standard ViT architecture. This strategy reduces the number of parameters in the model and results in a more lightweight model, without significantly compromising performance.

For the segmentation decoder, our three candidates are FANet~\cite{Hu2020RealTimeSS}, UPerNet~\cite{Xiao2018UnifiedPP} and DeepLabv3~\cite{Chen2017RethinkingAC}. FANet is a segmentation model that can perform real-time segmentation and it achieves this by using a feature aggregation network to produce high-level features, which are then used in a feature upsampling network for pixel-level predictions. UPerNet, on the other hand, enhances the resolution of the predictions by bringing together different levels of features using a multi-level feature fusion module and a pyramid pooling module. Finally, DeepLabv3 is a cutting-edge model that captures multiscale context using atrous convolution and incorporates low-level features using skip connections.

\begin{table}[!h]
    \caption{Experimental results for different backbones on semantic segmentation}
  \begin{center}  \begin{tabular}{|c|c|c|}
\hline
     Backbone & Params & Mean IoU \\
     \hline
     MobileViT & 13.4M & 37.06 \\
     \hline
     Swin-v2-T & 60M & 44.51 \\
     \hline
\end{tabular}\label{table-b1}
\end{center}
\end{table} 

Results of experiments conducted on ADE20K, a common benchmark for semantic segmentation, using the aforementioned networks as backbones have been reported in Table~\ref{table-b1}. Note that for a difficult dataset like ADE20K, a difference of 7 in mean IoU is considered very large, and we intend to deploy a MobileViT on the Jetson Nano with similar accuracy as Swin Transformer by applying our compression methods.



\section{Results and Analysis: LPCV Challenge Dataset}
\label{sec:results}
Given the large volume of existing network architectures for semantic segmentation, our first step is to identify a segmentation decoder architecture with the best trade-off between efficiency and accuracy on the LPCV dataset. In addition, by comparing the accuracy of different backbones, we utilize (1) a teacher network for distillation, and (2) our target backbone that we compress (pruning + quantization).

\subsection{Quantitative Evaluation}
The results are shown in Table~\ref{table-lpcv}. We include throughput (frames per second, FPS), accuracy, mean IoU, and model size in the study. For all results, we use an ImageNet-1k pretrained backbone and a randomly initialized decoder. We then finetune the models for 20,000 iterations on the LPCV dataset. 

As shown in Table~\ref{table-lpcv}, out of the three decoders - FANet, DeepLabv3 and UPerNet, FANet offers the least mean IoU with the fewest parameters, while DeepLabv3 offers the highest mean IoU with the largest model. We choose UPerNet, a middle ground between the two, to conduct our backbone comparison. Out of the four backbones we compared, we noticed that: (1) although ResNet18 has significantly more parameters than MobileViT, it has faster runtime. This is due to the self-attention matrix multiplication computation not contributing to parameter count. Also, convolution kernels in PyTorch seem to be more optimized than attention kernels. (2) Swin-v2-Tiny offers similar, although slightly higher IoU than Swin-v2-Base at just 40\% of the parameters. We believe that this is due to the LPCV dataset's small size. Swin-v2-Tiny is already capable of learning from data given our training configuration (data augmentation, etc.). 

\begin{table}[!h]
\centering
\caption{Comparison of different models on the LPCV Challenge Dataset}
\scalebox{0.8}{
    \begin{tabular}{|c|c|c|c|c|c|c|}
\toprule
\multicolumn{7}{|c|}{\textbf{Conventional Method}} \\
\hline
     Arch. + Backbone & FPS (GPU: RTX 4090) & FPS (Jetson Nano)
& Acc. & MIoU & \# Params & Score\\
     \midrule
     FANet (ResNet18) & 120.48 & - & 0.82 & 0.4752 & 13.9M & -\\
     \hline
     DeepLabv3 (MobileViT) & 64.93 & - & 0.85 & 0.5872 & 6.4M & -\\
     \hline
     UPerNet (ResNet18) & 76.49 & 6.02 & 0.84 & 0.5200 & 16.4M & 3.130
\\
     \hline
     UPerNet (MobileViT) & 56.75 & 3.11 & 0.85 & 0.5365 & 5.6M & 1.668 \\
     \hline
     UPerNet (Swin-v2-T) & 47.80 & - & 0.88 & 0.6468  & 40M & -\\
     \hline
     UPerNet (Swin-v2-B) & 33.70 & - & 0.88 & 0.6436 & 108M& -\\
     \bottomrule
\multicolumn{7}{|c|}{\textbf{Distillation-based}} \\
\hline
     UPerNet (MobileViT, KD) & 56.75 &3.11& 0.86 & 0.6056  & 5.6M&1.883 \\
     \hline
     UPerNet (ResNet18, KD) & 76.49 &6.02& 0.85 & 0.5493 & 16.4M&3.307\\
     \bottomrule
\multicolumn{7}{|c|}{\textbf{Distillation and Quantization}}\\
\hline
\textbf{UPerNet (MobileViT, KD, fp16)}& 57.07 & 4.49 & 0.86 & 0.6056 & 5.6M & 2.719\\
\hline
UPerNet (ResNet18, KD, fp16)& 81.75 & 5.94 & 0.85 & 0.5493 &16.4M &3.262 \\
\bottomrule
\end{tabular}}
\label{table-lpcv}
\end{table}

We choose UPerNet with Swin-v2-T as the teacher model and apply knowledge distillation with simple MSE Loss between the final segmentation logits. As shown in Table~\ref{table-lpcv}, UPerNet with MobileViT has a significant increase in IoU (0.06) after distillation, while ResNet18 + UPerNet sees a 0.02 increase. We also experiment with logit-based knowledge transfer using both simple MSE Loss and a KL-divergence loss of softmax of different temperatures. We found that consistently, the simple MSE Loss works better. On the other hand, using KL-divergence loss with temperatured softmax led to unstable training. Our final network after knowledge distillation: UPerNet with MobileViT, achieves 0.61 mean IoU while using only 5.6M parameters. 


In addition, we report the throughput of these models on a Jetson Nano. Consistent with FPS results on desktop GPUs, ResNet18 achieves much better runtime than MobileViT although having a bigger model size. Since we are limited by PyTorch's support of datatypes, we quantize the model to half-precision floats and report their FPS. Interestingly, while MobileViT saw a 1.4x increase in throughput, the throughput for ResNet18 at fp16 decreased. We believe this is still associated with PyTorch's implementation of convolution vs. attention kernels in CUDA.

In our final verification experiments, ResNet18 still achieves the best score (3.307). The best score achieved by a MobileViT is 2.719. However, MobileViT achieves a much higher mean IoU. Additionally, MobileViT still has a theoretical advantage in both model size and FLOPs, which implies that with the continuing optimization of software kernels for more recent computations like the attention mechanism, it has the potential of achieving better accuracy vs. runtime trade-off than conventional ConvNets. Our final model (distilled and quantized UPerNet + MobileViT) consumes 3742 MB of RAM with 1030 MB of swap on the Jetson Nano, while when idle, the OS alone consumes 1424+558 MB of RAM+swap. With optimized baremetal solutions and inference engines like ONNX or TensorRT, the memory footprint can be further optimized.


\subsection{Pruning Ablation}
\begin{figure}
    \centering
    \includegraphics[width=\textwidth]{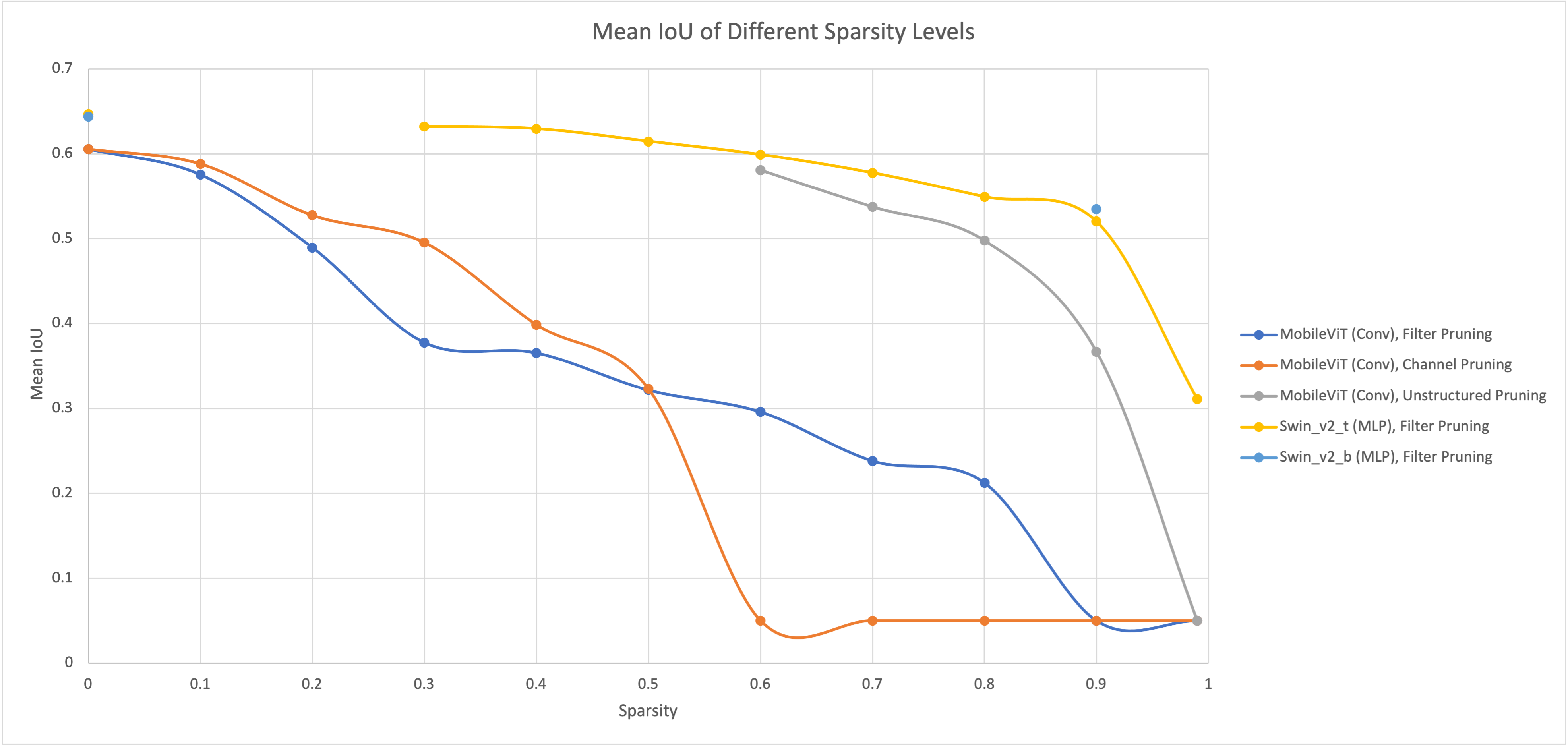}
    \caption{\textbf{Effects of Pruning Convolutional and Linear layers:} MIoU vs Sparsity trends for different models}
    \label{fig-convprune}
\end{figure}

Our first pruning attempt is filter pruning for the Conv layers in MobileViT. Our initial estimation is to get at least 0.5 sparsity since traditional ConvNets with similar size (MobileNet) are able to achieve that with negligible loss in accuracy. The results are shown in Figure~\ref{fig-convprune}. As observed in the plot, starting at 0.2 sparsity results in a significant drop in mean IoU.

Our conjecture is that MobileViT is already compressed to a high degree, in order to be able to tolerate any additional pruning. In order to understand this hypothesis further, we attempt channel pruning, unstructured pruning on MobileViT, as well as filter pruning on the much larger Swin-v2-T and Swin-B models. As shown in Figure~\ref{fig-convprune}, channel pruning results in a similar accuracy degradation compared to filter pruning, while unstructured pruning can achieve much high sparsity. When pruning Swin models, accuracy only starts to degrade at around 0.7 sparsity, corroborating our hypothesis about MobileViT's inability to further compress. We note that Swin Transformers contain MLP layers instead of Conv layers, which have more inherent redundancy than MobileViT.

\begin{table}[!h]
\centering
\caption{Effects of head pruning on different models}
\scalebox{0.8}{
    \begin{tabular}{|c|c|c|c|c|}
    \hline 
    Model & \# Attention Heads & Mean IoU & \# Params & FPS (GPU)\\
    \hline 
    UPerNet (MobileViT) & 4 & 0.545 & 5.6M & 62\\
    \hline 
    UPerNet (MobileViT) & 2 & 0.533 & 4.6M & 62\\
    \hline 
    DeepLabv3 (MobileViT) & 4 & 0.601 & 6.4M & 64\\
    \hline 
    DeepLabv3 (MobileViT) & 2 & 0.590 & 5.4M & 64\\
    \hline
    \end{tabular}
}
\label{table-headprune}
\end{table}

Next, we perform pruning of attention heads based on L2 norm of the weights on the models with transformer-based backbones. The results are shown in Table \ref{table-headprune}. The results show that reducing the number of heads from 4 (the default value for MobileViT) to 2 had minimal impact on model performance while offering a significant reduction in the number of parameters. Of note, we pruned the weights by setting the selected heads to zero but did not actually change the dimensions of the attention block. Therefore, the parameter reduction is theoretical, and we do not see an actual improvement in speed as the number of operations does not change. 

\subsection{Qualitative Evaluation}
\begin{figure*}[!h]
    \centering
    \includegraphics[width=0.8\textwidth]{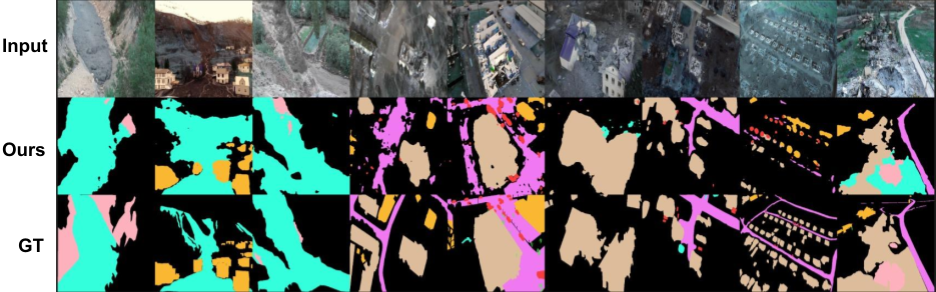}
    \caption{Segmentation results on the LPCV dataset}
    \label{fig-qualcomp}
\end{figure*}

Using the UPerNet with MobileViT distilled from Swin-v2-T and fp16 quantization as our final framework, the segmentation results on the LPCV dataset are shown in Figure \ref{fig-qualcomp}. We can see from the resulting images that our compressed model actually does a fairly good job at capturing the dominant features of the image, but fails to accurately capture finer details. For example, on the second image from the right, our compressed model has difficulty identifying distant roads. This task is particularly difficult as many classes look very similar, even to the human eye. The differences between mudflow and a dirt road are subtle and often not apparent.  

This issue becomes further complicated as we have been evaluating our models on global performance. Our compression techniques are only preserving the parameters and interactions that are the most performant at a global scale. Therefore, it is reasonable to expect that not all classes would be impacted in a similar manner as classwise distributions are, by the nature of disaster scenes, not equal. This is evident even in our example images as only a few classes, including the background, constitute the overwhelming majority of the pixels. This issue of classwise imbalance becomes problematic because, in disaster scene parsing, the low-frequency classes are actually the most important in a real-world scenario.

\subsection{Unsuccessful Experiments: Iterative Pruning}
\begin{figure}[!h]
    \centering
    \includegraphics[width=0.7\textwidth]{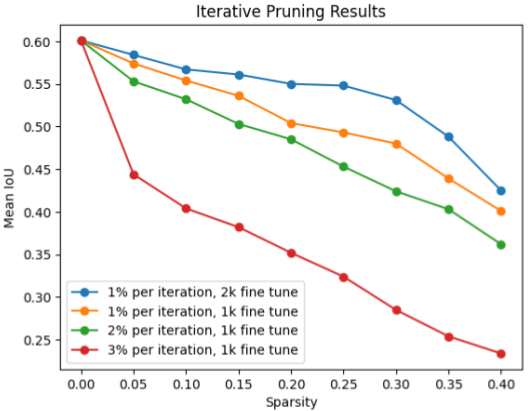}
    \caption{MIoU vs Sparsity trends for iterative pruning}
    \label{fig-iterprune}
\end{figure}

We perform iterative pruning in order to try and achieve better pruning results on MobileViT as shown in Figure \ref{fig-iterprune}. Our approach involves initially pruning a small percentage of filters remaining in each convolution layer, followed by 1k or 2k fine-tuning iterations. We then repeat this cycle and collect performance metrics for comparison. Our hypothesis was that by not removing many filters at once, the model may be able to better preserve performance by redistributing the weights during the fine-tuning iterations. We find that this process gives marginal improvements over the bulk pruning described previously, but still does not provide much avenue for compression. 


\section{Summary}

In our work, we establish baseline ConvNet and Transformer models for segmentation on the LPCV dataset. We devise a framework to compress these models to meet the memory usage requirements of no more than 4GB of RAM system-wide. Our compression approach builds on structured pruning to remove redundant heads from the transformer layers, along with linear layer weights. Furthermore, we also leverage the higher expressive power of a stronger teacher network using distillation, which guides our smaller, more efficient student network. In order to strike the optimal balance between memory usage and inference latency improvements while maintaining segmentation accuracy and mean IoU, we further explore quantization strategies and employ the most effective method on the student network at inference time on the Jetson Nano. We also explore other compression techniques and their effectiveness in reducing the model size and inference latency. On the whole, our work focuses on and successfully develops a framework to enable highly efficient segmentation on the edge that could be deployed on resource-constrained devices for real-world applications.


\section{Future Directions}

Moving forward, our future work will focus on further compression and evaluation of our model on the low-cost NVIDIA Jetson Nano \textbf{2GB}. This environment provides the right balance of low power, affordability, and robust support in a remote execution setting, especially in disaster-stricken regions.

We also plan to experiment and analyze other quantization techniques and their effectiveness in reducing inference latency while remaining performant. Additionally, we will investigate the limitations of PyTorch's support for quantized CUDA operations and seek alternative solutions if necessary. Beyond this, we hope to explore low-rank approximation of the attention mechanism for our use case.

 We also aim to explore adaptive pruning methods such as Layer-Adaptive Magnitude-based Pruning (LAMP)~\cite{Lee2020LayeradaptiveSF} to achieve a better trade-off between sparsity and performance without requiring extensive hyperparameter tuning or heavy computation.

\bibliographystyle{unsrt}
\bibliography{neurips_2023}






\end{document}